\documentclass[runningheads]{llncs}
\usepackage{graphicx}
\usepackage{booktabs}
\usepackage{wrapfig}
\usepackage{multirow}
\usepackage{floatrow}
\usepackage{subcaption}
\usepackage[linesnumbered,ruled,vlined]{algorithm2e}

\SetCommentSty{mycommfont}
\newcommand\difffont[1]{{\textcolor{red}{#1}}}

\SetKwInput{KwInput}{Input}                
\SetKwInput{KwOutput}{Output}              
\usepackage{mathrsfs}
\usepackage{tikz}
\usepackage{comment}
\usepackage{amsmath,amssymb} 
\usepackage{color}

\usepackage{makecell}

\newfloatcommand{figurebox}{figure}[\nocapbeside][\dimexpr(\textwidth-\columnsep)/2\relax]
\newfloatcommand{tablebox}{table}[\nocapbeside][\dimexpr(\textwidth-\columnsep)/2\relax]

\usepackage{xspace}
\def\onedot{.\xspace}

\def\eg{\emph{e.g}\onedot} 
\def\ie{\emph{i.e}\onedot}

\def\modelName{RegionCL\xspace}

\usepackage[accsupp]{axessibility}  

\begin{document}
\pagestyle{headings}
\mainmatter
\def\ECCVSubNumber{1165}  

\title{
RegionCL: Exploring Contrastive Region Pairs for Self-supervised Representation Learning
} 

\titlerunning{RegionCL}
%
\author{Yufei Xu\inst{1}\thanks{Equal contribution.} \and
Qiming Zhang\inst{1}* \and
Jing Zhang\inst{1} \and 
Dacheng Tao\inst{2,1}}
\authorrunning{Y. Xu et al.}
%
\institute{University of Sydney, Australia \and
JD Explore Academy, China \\
\email{\{yuxu7116,qzha2506\}@uni.sydney.edu.au, \\ jing.zhang1@sydney.edu.au, dacheng.tao@gmail.com}}
\maketitle
\begin{abstract}

Self-supervised learning methods (SSL) have achieved significant success via maximizing the mutual information between two augmented views, where cropping is a popular augmentation technique. Cropped regions are widely used to construct positive pairs, while the remained regions after cropping have rarely been explored in existing methods, although they together constitute the same image instance and both contribute to the description of the category. In this paper, we make the first attempt to demonstrate the importance of both regions in cropping from a complete perspective and {the effectiveness of using both regions via designing a simple yet effective pretext task called Region Contrastive Learning (\modelName). Technically, to construct the two kinds of regions, we randomly crop a region (called the paste view) from each input image with the same size and swap them between different images to compose new images together with the remained regions (called the canvas view). Then, instead of taking the new images as a whole for positive or negative samples, contrastive pairs are efficiently constructed from the regional perspective based on the following simple criteria, i.e., each view is (1) positive with views augmented from the same original image and (2) negative with views augmented from other images.} With minor modifications to popular SSL methods, \modelName exploits those abundant pairs and helps the model distinguish the regions features from both canvas and paste views, therefore learning better visual representations. Experiments on ImageNet, MS COCO, and Cityscapes demonstrate that \modelName improves MoCov2, DenseCL, and SimSiam by large margins and achieves state-of-the-art performance on classification, detection, and segmentation tasks. The code is publicly available at \href{https://github.com/Annbless/RegionCL}{https://github.com/Annbless/RegionCL}.

\end{abstract}

\section{Introduction}

Self-supervised learning (SSL) has become an active research topic in computer vision because of its ability to learn generalizable representations from large-scale unlabeled data and offer good performance in downstream tasks~\cite{zhang2020empowering,zhang2022vitaev2,li2022exploring,chen2017deeplab,wang2020self}. Contrastive learning, one of the popular directions in SSL, has attracted a lot of attention due to its ease of use in pretext designing and capacity to generalize across various visual tasks.

Current contrastive learning methods typically use augmented views of the same image as positive pairs and maximize their mutual information. Cropping is by far the most popular augmentation technique. By randomly cropping regions from the same images and treating the cropped regions as positive pairs, the methods in \cite{chen2020a,chen2020improved,chen2021exploring,grill2020bootstrap,he2020momentum,chen2020big} have shown promising results in image classification. Multi-crop~\cite{caron2020unsupervised,caron2021emerging} has been investigated as a way to improve performance even further by generating more diverse candidates and facilitating the model learning a better feature representation. Constrained cropping strategies~\cite{zhao2021self,xie2021detco,yang2021instance,roh2021spatially,xie2021propagate} have recently been developed to ensure that two cropped views contain shared regions of a specific size and to improve models' performance on dense prediction tasks by constructing contrastive pairs within the shared regions. These methods have achieved superior performance on a variety of visual tasks by leveraging various cropping strategies to construct contrastive pairs during pretraining. 

However, the remained regions after cropping have received little attention, despite the fact that the cropped and remained regions together make up the same image instance and both contribute to the category's description.
We argue that using both regions during pretraining would help the model learn better complete visual representations of object instances, which will improve the model's performance on downstream classification and dense prediction tasks.

\thisfloatsetup{heightadjust=all,valign=c}
\begin{figure}[t!]
\begin{floatrow}[2]
\figurebox{
  \caption{Transfer results on the ImageNet~\cite{deng2009imagenet} (classification) and MS COCO~\cite{chen2015microsoft} (detection) datasets. Involving region-level contrastive pairs during pretraining, \modelName helps DenseCL~\cite{wang2021dense}, MoCov2~\cite{chen2020improved}, and SimSiam~\cite{chen2021exploring} achieve a better performance trade-off between image classification and object detection tasks.}
  \label{fig:opening}
  }
  {%
  \includegraphics[width=\linewidth]{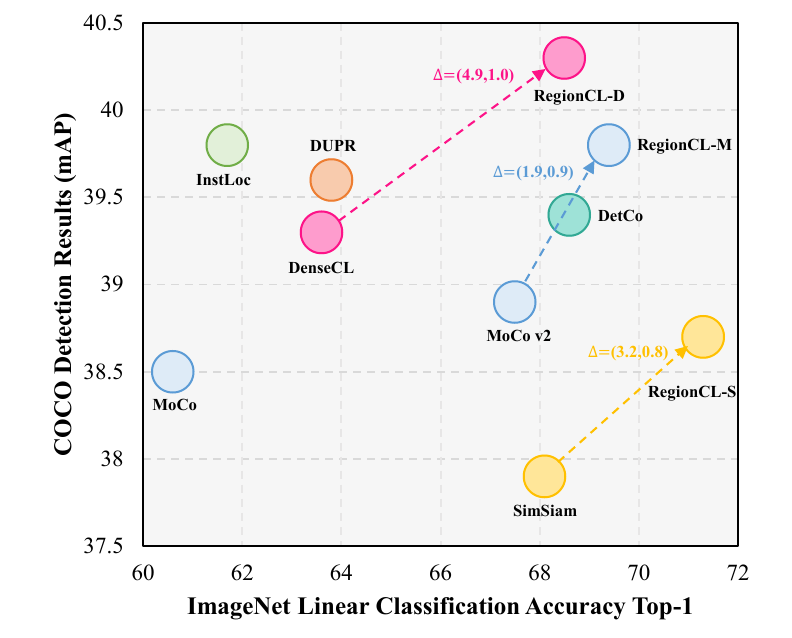}
  }%
\figurebox{
    \caption{Illustration of the region swapping strategy. Taking two images A and B as input, it randomly crops and swaps the paste views and generates the composite images C and D. As a result, the canvas and paste views in image C form a negative pair, while the canvas view and A (the paste view and B) are positive pairs.}
    \label{fig:motivation}
    }
  {%
  \centering
  \includegraphics[width=1.\linewidth]{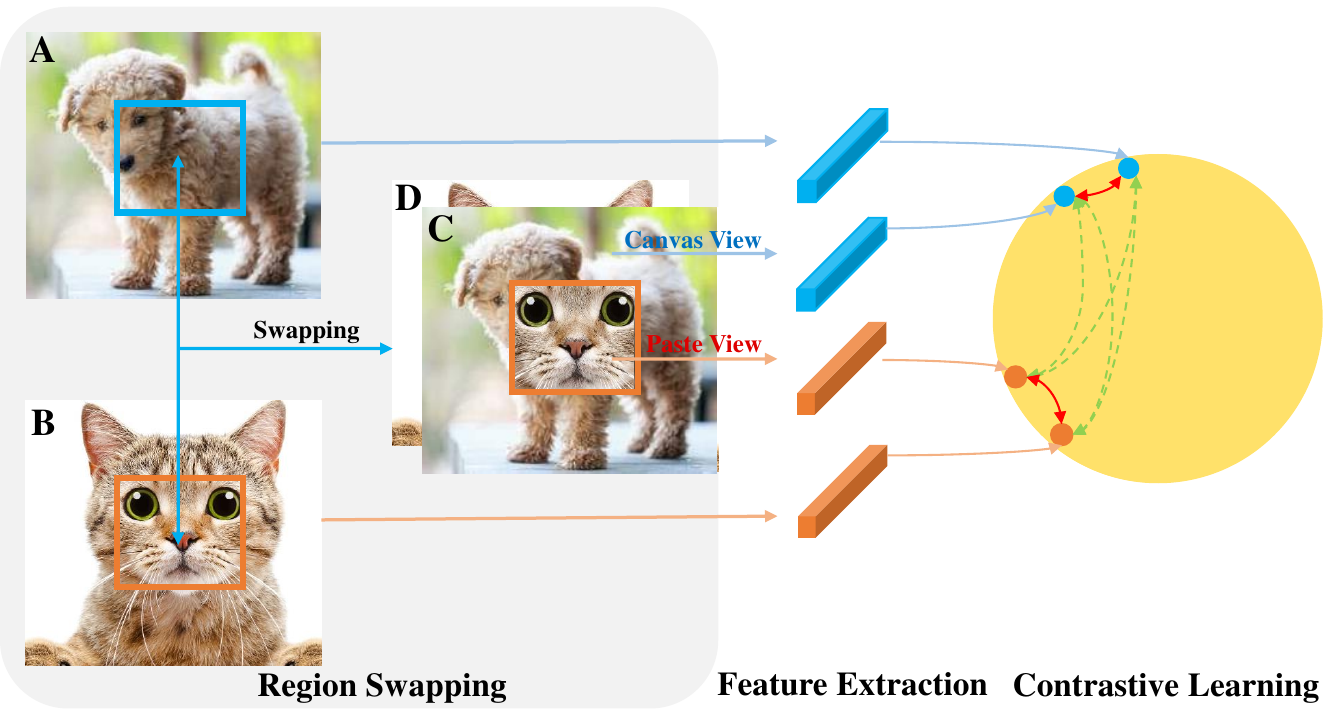}
  }
\end{floatrow}
\end{figure}

{Based on this motivation, we propose a simple yet effective pretext task called Region Contrastive Learning (\modelName) to demonstrate the effectiveness of using both regions for contrastive learning.} Technically, given two different images, \modelName randomly crops a region (called the \textbf{paste view}) from each image with the same size and swaps them to compose two new images together with the remained regions (called the \textbf{canvas view}), respectively. It is worth noting that the two views that compose the new images are from different source images. Then, contrastive pairs can be constructed from the regional perspective following the simple criteria, \ie, each view is (1) positive with views augmented from the same original image and (2) negative with views augmented from other images. In this way, \modelName generates abundant pairs that contain not only the instance-level pairs as other methods~\cite{chen2020improved,chen2020a} but also the region-level pairs, \eg, the paste and canvas views in the composite images. By exploiting these pairs in popular SSL frameworks, \modelName helps the models learn better feature representations of object instances owing to the abundant contrastive supervisory signals at both instance and region levels, delivering better performance on various downstream tasks. As shown in Figure~\ref{fig:opening}, \modelName helps MoCov2~\cite{chen2020improved}, DenseCL~\cite{wang2021dense}, and SimSiam~\cite{chen2021exploring} improve their linear classification accuracy by 2\%$\sim$5\% on the ImageNet~\cite{deng2009imagenet} dataset and object detection performance by 0.8$\sim$1.0 mAP on the MS COCO~\cite{chen2015microsoft} dataset, simultaneously.

In summary, the contribution of the paper is threefold:
\begin{enumerate}
    \item We make the first attempt to demonstrate the importance of both regions, \ie, the cropped and remained regions in cropping, from a complete perspective for self-supervised learning.
    \item We propose a simple yet effective pretext task, \ie, \modelName, {to demonstrate the effectiveness of using both regions for learning.} It is compatible with various popular SSL methods with minor modifications and improves their performance on many downstream visual tasks.
    \item Extensive experimental results with MoCov2, SimSiam, and DenseCL on the ImageNet, MS COCO, and Cityscapes datasets demonstrate the effectiveness of the proposed \modelName on classification, detection, and instance and semantic segmentation tasks. 
\end{enumerate}

\section{Related Work}

Self-supervised learning has shown great potential in learning visual representations that can generalize to a series of downstream visual tasks. Early works generate pseudo labels using specific tasks~\cite{noroozi2016unsupervised,larsson2016learning,zhang2016colorful,pathak2016context} such as image corruption and restoration, reordering, re-colorization. However, the models pretrained in these tasks may be too coupled with the designed tasks and the transfer results on other visual tasks may not be competitive.

Recently, contrastive learning~\cite{chen2021exploring,chen2020improved,chen2020a,he2020momentum,grill2020bootstrap,tian2020contrastive,chen2020big} has made rapid progress and shown promising transfer performance. Typically, they take augmented views from the same (different) images as positive (negative) pairs and learn to pull the features from positive pairs while pushing away those from the negative pairs via a contrastive loss. Among the augmentation techniques, cropping plays an important role in improving the performance, as shown in \cite{chen2020a}. Taking the cropped augmented views as input, SimCLR~\cite{chen2020a} obtains superior results on image classification. MoCo~\cite{he2020momentum,chen2020improved} utilizes a momentum encoder to better utilize the cropped views during pretraining, as it provides consistent optimization direction. However, as cropping at a single resolution may not provide enough descriptions of the target object, a multi-crop strategy is explored in \cite{caron2020unsupervised,caron2021emerging} by fusing several cropped views at different resolutions. Such a strategy helps the models learn a better feature representation at different scales and boost their performance on the image classification task. 

On the other hand, \cite{wang2021dense,xie2021propagate,roh2021spatially,pinheiro2020unsupervised} focus on advancing the performance on dense prediction tasks by establishing dense correspondences between the augmented cropped views. Some methods~\cite{zhao2021self,roh2021spatially,chen2021multisiam,liu2020selfemb} further design contrained cropping strategies during pretraining to improve the transfer results on detection, \eg, they require the two cropped views have some shared regions and attract the dense positive features within the shared regions based on explicit spatial correspondences.
By exploring different properties of cropping-based augmented views, these methods obtain superior performance. However, the remained regions after cropping have rarely been explored. Different from them, we make the first attempt to investigate the importance of both regions in this study {via adopting a simple region swapping strategy to generate abundant contrastive pairs at both instance and region levels for contrastive learning (\modelName), from which the model can learn better visual representations of object instances.}

Although several methods also explore region-level contrastive learning, they have not yet explored the complementary remained regions after cropping, \ie, the canvas view in our paper. For example, SCRL~\cite{roh2021spatially}, DUPR~\cite{ding2021unsupervised}, and MaskCo~\cite{zhao2021self} incorporate bounding boxes generation and alignment between the shared area of two cropped views during pretraining. InstLoc~\cite{yang2021instance} further introduces anchors with bounding boxes augmentations to boost the transfer results on dense prediction tasks at the cost of decreased image classification accuracy. DetCo~\cite{xie2021detco} designs delicate cropping strategies to generate separate patches at different resolutions and uses extra memory banks to capture patch features. 
{Without the requirement of extra information such as bounding boxes alignment, \modelName adopts the most simple strategy to demonstrate that using both regions for contrastive learning isolated helps to boost the SSL methods performance without bells and whistles. The simple task is compatible with popular SSL frameworks, \eg, we validate the effectiveness of using both regions for training with MoCov2~\cite{chen2020improved}, SimSiam~\cite{chen2021exploring}, and DenseCL~\cite{wang2021dense}, with only minor modifications to them. The theoretical analysis of these works have been well studied in \cite{tian2022deep,tao2021exploring,wang2021understanding,wang2020understanding} and is beyond this paper's scope. It is also noteworthy that although the adopted swapping strategy is similar to co-current methods like UnMix~\cite{UnMix}/HEXA~\cite{HEXA}/InsCon~\cite{yang2022inscon}, they do not explore region-level pairs, \ie, they still treat the composite images as a whole from the global perspective. \modelName explores both region- and global-level pairs from the composite images, which is more efficient and obtains better performance on various vision tasks.} 

\section{Method}

\begin{figure}[t!]
    \centering
    \includegraphics[width=1.\linewidth]{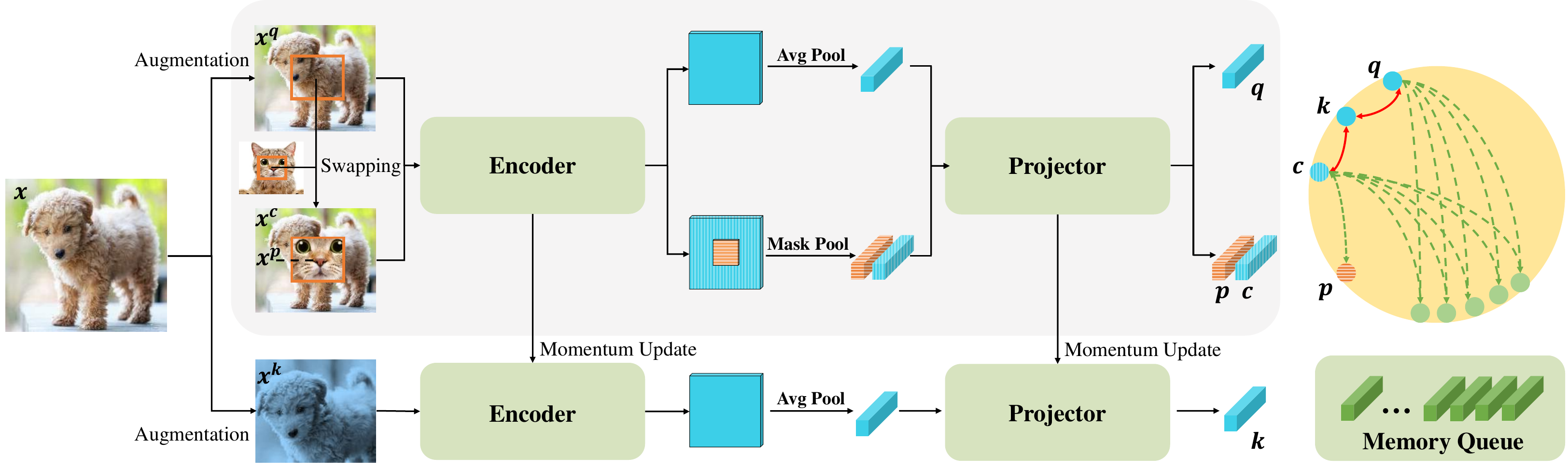}
    \vspace{-3 mm}
    \caption{Illustration of the proposed \modelName with the MoCov2 framework, \ie, \modelName-M. Taking the two augmented views $x^q$, $x^k$ as inputs, \modelName employs region swapping among the batch of $x^q$ to generate the composite images with paste views $x^p$ and canvas views $x^c$. Then, mask pooling is used to extract the features belonging to the paste and canvas views, respectively. The resultant region-level features (with stripes in the figure) are batched with the instance-level features and processed by the projector.}
    \label{fig:pipeline}
\end{figure}

\subsection{The region swapping strategy}
\label{subsec:regionswapping}
Different from current methods that only use the cropped regions, we take both the cropped and remained regions into consideration for self-supervised learning. Given two different images, we randomly crop a region with the same size from each image and swap them to compose two new images. As shown in Figure~\ref{fig:motivation}-C, the composite image after region swapping contains two views: one is the paste view (the cropped region), \ie, the cat's face, and the other is called the canvas view (the remained region), \ie, the dog's body. Specifically, we first sample a size of the paste view, \ie, the height and width, and then determine the coordinates of the origin point from which the cropping starts. We make sure the size and location of the cropped region match the network's downsampling ratio $\mathcal{R}$ during region swapping so that the region's feature can be directly extracted from the feature map by a simple operation of mask pooling. The height and width are determined by $\mathcal{R}$ and a discrete uniform distribution $\mathcal{C} \sim U(\mathcal{C}_L, \mathcal{C}_U)$, where the ratio $\mathcal{R}$ is typically $32$ for ResNet~\cite{he2016deep} and $\mathcal{C}_L,\mathcal{C}_U$ are two predefined hyper-parameters shared for both spatial dimensions for simplicity. They are set to 3 and 5 in the paper unless specified. We sample twice from the distribution $\mathcal{C}$ and get two observations $c_h$ and $c_w$. Then, we calculate the width and height as $r_h=c_h \times \mathcal{R}, r_w=c_w \times \mathcal{R}$, respectively. Then we uniformly sample the origin point coordinates ($r_x,r_y$) from a valid range that guarantees there is enough remaining area to crop a patch of size $r_w \times r_h$. In this way, the candidate region is determined by $(r_x, r_y, r_w, r_h)$. 
It is noteworthy that within a mini-batch of the training images, we use the same coordinates $(r_x, r_y, r_w, r_h)$ for efficient batch-wise implementation during training.

\subsection{The region contrastive learning}
\noindent \textbf{The architecture.}
We take MoCov2 as an example here to describe the proposed \modelName method in depth, denoted as \modelName-M. The overall architecture is presented in Figure~\ref{fig:pipeline}. As can be seen, \modelName-M has exactly the same architecture with MoCov2 and only requires marginal modifications to the inputs and learning objectives, \ie, a region-level branch in the middle.

\modelName-M uses a Siamese network structure in pretraining. Given image instances $x$, \modelName-M first creates two randomly augmented views, \ie, the query view $x^q$ and the key view $x^k$ following the same augmentation strategy as in MoCov2. The online network processes the query view, and the other branch, \ie, the momentum updated network, processes the key view. Unlike other methods that also utilize region-level contrastive learning \cite{roh2021spatially,ding2021unsupervised,zhao2021self}, we follow the same cropping strategies as in MoCov2 and do not need the two views $x^q,x^k$ to have a sufficiently large overlap, which keeps the diversity of the contrastive pair candidates. We construct the region-level contrastive pairs using the region-swapping strategy. Specifically, given two image instances from the query view $x^q$, we randomly crop a region with the same size in each image instance and swap them to compose two new images $x^{pc}$, where the cropped region after swapping and the remained region in the new images are the paste view $x^p$ and canvas view $x^c$, respectively.

\noindent \textbf{The region- and instance-level contrastive loss.}
In this way, we have a total of four different views, \ie, the query view, the paste view, the canvas view, and the key view, denoted as $x^q$, $x^p$, $x^c$, $x^k$, respectively. \modelName-M projects these views into the corresponding feature representations $q$, $p$, $c$, $k$, among which the features $q$, $k$ are instance-level feature representations while the features $p,c$ are region-level feature representations. Note that features of the paste view and canvas view are extracted from the feature maps of $x^{pc}$ via mask pooling, where the mask is obtained according to the coordinates $(r_x, r_y, r_w, r_h)$ as described in Section~\ref{subsec:regionswapping}. The other views' features are from the global average pooling upon the corresponding feature maps. Then we can efficiently construct the contrastive pairs for these views according to the simple criteria, \ie, each view is (1) positive with views augmented from the same original image and (2) negative with views augmented from other images. We follow the practice of MoCov2 in our implementation and ignore the positive pairs whose features are both generated by the online network to stabilize the training.

We use contrastive loss~\cite{hadsell2006dimensionality,he2020momentum} as the learning objectives, which can be thought of as training an encoder for a dictionary lookup task at both instance and region levels. We first introduce the instance-level contrastive loss and then present the region-level one. Assume that we have a set of encoded samples $\{ k_i | i=1,2,...,K\}$ as keys of a dictionary. For each query feature $q$, if there is a single key ($k^{+}$) that matches the query $q$, the contrastive loss aims to increase the similarity between $q$ and $k^+$ meanwhile reducing the similarity between $q$ and all other keys (considered as the negative counterparts for $q$). We use L2-normalized dot product to measure the similarity between the queries and keys, and the contrastive loss, \ie, the InfoNCE~\cite{oord2018representation} loss, is therefore formulated as:
\begin{equation}
    \mathcal{L}_{ins} = - \log \frac{\exp{(q \cdot k^+ / \tau)}}{\sum_{i=0}^K \exp{ (q \cdot k_i / \tau) }},
    \label{eq:Lq}
\end{equation}
where $\tau$ is a temperature hyper-parameter (set to 0.2 by default)~\cite{wu2018unsupervised,he2020momentum}. Following MoCov2, the dictionary keys $\{ k_i | i=1,2,...,K \}$ in \modelName-M are maintained using a first-in-first-out queue with a predefined maximum number of samples ($K$), which is set to 65,536. The features from the key view $k$ is treated as the positive sample and used to progressively update the memory queue, which serves as the negative samples. This form of contrastive loss is the exact one that appeared in MoCov2~\cite{he2020momentum}, while it can have other forms for different SSL methods \cite{oord2018representation,chen2021exploring}. Apart from the instance-level pairs, the features of other views formulate the region-level pairs with the modified contrastive loss:
\begin{equation}
\begin{aligned}
    \mathcal{L}_{reg} = & - \frac{1}{2} \log \frac{\exp{(p \cdot k^+_p / \tau)}}{\sum_{i=0}^K \exp{ (p \cdot k_i / \tau) + \exp{(p \cdot sg(c)} / \tau) }} \\
    & - \frac{1}{2} \log \frac{\exp{(c \cdot k^+_c / \tau)}}{\sum_{i=0}^K \exp{ (c \cdot k_i / \tau) + \exp{(c \cdot sg(p) / \tau} )}},
    \label{eq:Lp}
\end{aligned}
\end{equation}
where the features $p,c$ are obtained from the identical composite image $x^{pc}$ (thus the term is divided by $\frac{1}{2}$ for normalization). $sg(\cdot)$ represents `stop gradient', which helps stabilize the training. $p$ and $c$ are indeed hard negative pairs since they involve some context information from each other due to convolution and pooling operations, thereby helping the model to learn robust and discriminative feature representations.
Thus the total contrastive loss is formulated as:
\begin{equation}
    \mathcal{L}_{total} = \mathcal{L}_{ins} + \mathcal{L}_{reg}.
\end{equation}
 Since the query, canvas, and paste views share the online network, we believe that the features of these views should be in the same feature space. Thus, \modelName-M only needs a single queue to provide negative samples for features of all the three views, in contrast to the usage of multiple queues as in \cite{xie2021detco,yang2021instance}. 

\subsubsection{Extension to other SSL methods.}

As \modelName defines a model-agnostic pretext task and requires minor modifications to the SSL methods, we also choose two other representative approaches, \ie, DenseCL~\cite{wang2021dense} and SimSiam~\cite{chen2021exploring}, to further validate its effectiveness, denoted as \modelName-D and \modelName-S, respectively. Specifically, DenseCL~\cite{wang2021dense} focuses on dense prediction tasks and has two learning objectives, \ie, the instance-level contrastive loss as in MoCov2 and the pixel-level dense loss. Therefore, \modelName-D includes the proposed region-level loss seamlessly as in \modelName-M and keeps the pixel-level loss unchanged. For SimSiam~\cite{chen2021exploring}, it only adopts instance-level positive pairs $\{p,k^+\}$ for training. Thus, we only enrich the positive pairs by collecting those abundant instance- and region-level positive pairs provided by \modelName-S while retaining the other components. Similar strategy is adopted for vision transformer backbones~\cite{xu2021vitae,liu2021swin,dosovitskiy2020image} with MoBy~\cite{MOBY}. Please refer to the supplementary for details.

\section{Experiments}

To thoroughly validate the improvements brought by introducing both regions into pretraining, we incorporate the \modelName in representative state-of-the-art SSL methods, \ie, MoCov2~\cite{chen2020improved}, DenseCL~\cite{wang2021dense}, and SimSiam~\cite{chen2021exploring}, and propose the \modelName compatible models, \ie, \modelName-M, \modelName-D, and \modelName-S. The models are pretrained following the same settings as their own base methods, \ie, we train \modelName-M and \modelName-D for 200 epochs, and \modelName-S for 100 epochs, with an SGD~\cite{sutskever2013importance} optimizer and corresponding augmentations, respectively. All the methods are based on ResNet-50~\cite{he2016deep} backbone. Please refer to the supplementary material for more details. {Vision transformer-based methods~\cite{MOBY} are also explored to further evaluate RegionCL's effectiveness.} 

\subsection{Image classification on ImageNet}

\noindent \textbf{Settings.} {To evaluate the effectiveness of regional contrastive learning for image classification,} we benchmark the \modelName on ImageNet~\cite{deng2009imagenet}, which contains 1.28M images in the training set and 50K images in the validation set from 1,000 classes, respectively. The pretrained models of other SSL methods are either obtained from their authors or reproduced using their official codes. The performance of Top-1 and Top-5 accuracy on a single crop is reported. We have two experimental settings regards evaluation: linear classification and few-shot finetuning. The former setting follows the default setting of MoCov2~\cite{he2020momentum,chen2020improved} and SimSiam~\cite{chen2021exploring} with SGD~\cite{sutskever2013importance} and LARS~\cite{you2017large} optimizer. The latter one using randomly sampled data per class from the training set.

\noindent \textbf{Results with linear classification.} We report the linear classification results of different methods in Table~\ref{tab:CLS} and \ref{tab:CLSLonger}. `Real' indicates that the labels used for evaluation are provided by \cite{beyer2020we}. From the table, we can see that \modelName improves the aforementioned SSL baseline methods significantly by a large margin: +1.9\% for \modelName-M, +4.8\% for \modelName-D, and +3.2\% for \modelName-S. 
{This proves \modelName helps various SSL methods learn better feature representations owing to the abundant contrastive supervisory signals with contrastive pairs from both regional and global perspective. Comparing with methods that also exploring mixing two images for contrastive learning but from the global perspective only, \ie, UnMix~\cite{UnMix} and HEXA~\cite{HEXA}, \modelName-M obtains better performance no matter using 200/800 epochs for training, demonstrating the benefits of leveraging regional contrastive pairs.}
Besides, \modelName-S reaches 71.3\% Top-1 accuracy using only 100 epochs, while the vanilla SimSiam requires a significantly longer training schedule of 800 epochs, proving the effectiveness of \modelName in accelerating the model convergence and improving the performance. It is noteworthy that DenseCL focuses on dense prediction tasks and does not perform that well on classification. In contrast, \modelName brings a large improvement on DenseCL for image classification, 
{indicating that {introducing remained regions to promote pretraining} is not only compatible with classification-favored SSL methods but also generalizes well on dense prediction-favored approaches.}

\noindent \textbf{Results with linear few-shot finetuning.} Table~\ref{tab:CLSFew} presents the results of different methods at the linear few-shot finetuning setting. Thanks to the abundant contrastive pairs brought by \modelName, the models pretrained by \modelName have learned better feature representations from a complete perspective and can generalize well on classification tasks, thus delivering much more significant improvements over their baselines when only a limited number of data are available for finetuning, \ie, a gain of +2.5\%, +8.9\%, +9.5\% for 1\% data and 1.8\%, 6.4\%, 8.1\% for 10\% data achieved by \modelName-M, \modelName-D, \modelName-S. 

\begin{table}\BottomFloatBoxes
    \begin{floatrow}
    \hsize1.2\hsize 
    \tablebox
    {\caption{Linear classification results comparison on ImageNet~\cite{deng2009imagenet}.}
    \label{tab:CLS}}
    {
    {
  \setlength{\tabcolsep}{0.005\linewidth}{
  \begin{tabular}{l|c|cc|c}
    \hline
          & \multirow{2}[1]{*}{Epochs} & \multicolumn{2}{c|}{ImageNet} & {Real} \\
          &       & Top-1 & Top-5 & Top-1\\
    \hline
    MoCo~\cite{he2020momentum} & 200   & 60.6  & - & 69.1 \\
    SimCLR~\cite{chen2020a} & 1000  & 69.3  & 89.0 & 77.6 \\
    PIRL~\cite{misra2020self} & 800 & 64.3 & - & 71.7\\
    CPC-v2~\cite{CPC} & 200 & 63.8 & 85.3 & - \\
    InstLoc~\cite{yang2021instance} & 200   & 61.7  & -  & - \\
    MaskCo~\cite{zhao2021self} & 200   & 65.1  & - & - \\
    ISD~\cite{ISD} & 200 & 69.8 & - & - \\
    PCLv1~\cite{PCL} & 200 & 61.5 & - & - \\
    PCLv2~\cite{PCL} & 200 & 67.6 & - & - \\
    DUPR~\cite{ding2021unsupervised} & 200   & 63.8  & 85.6 & - \\
    DetCo~\cite{xie2021detco} & 200   & 68.6  & 88.5 & - \\
    UnMix~\cite{UnMix} & 200 & 68.6 & - & - \\
    HEXA~\cite{HEXA} & 200 & 68.9 & - & - \\
    MoCov3~\cite{MoCov3} & 100 & 68.9 & - & - \\
    SimSiam~\cite{chen2021exploring} & 800   & 71.3  & - & - \\
    \hline
    MoCov2~\cite{chen2020improved} & 200   & 67.5  & 88.2 & 77.8 \\
    \textbf{\modelName-M} & \textbf{200} & \textbf{69.4} & \textbf{89.6} & \textbf{78.7} \\
    \hline
    DenseCL~\cite{wang2021dense} & 200   & 63.6  & 85.5  & 72.3 \\
    \textbf{\modelName-D} & \textbf{200}  & \textbf{68.5} & \textbf{89.0} & \textbf{78.4} \\
    \hline
    SimSiam~\cite{chen2021exploring} & 100   & 68.1  & 88.2 & 77.8 \\
    \textbf{\modelName-S} & \textbf{100} & \textbf{71.3} & \textbf{90.4} & \textbf{80.8} \\
    \hline
    \end{tabular}
    }
}
    }
    \vbox to10.1cm
    {
    \floatsetup[table]{floatrowsep=none}\killfloatstyle
    \tablebox
    {
    \caption{Linear classification results with more pretraining epochs. * means using symmetric loss.}
    \label{tab:CLSLonger}}
    {
  \setlength{\tabcolsep}{0.01\linewidth}{
    \begin{tabular}{l|c|cc}
    \hline
          & \multirow{2}[2]{*}{Epochs} & \multicolumn{2}{c}{ImageNet} \\
          &       & Top-1 & Top-5 \\
    \hline
    MoCov2~\cite{chen2020improved} & 800   & 71.1  & 90.2 \\
    UnMix~\cite{UnMix} & 800   & 71.8  & - \\
    HEXA~\cite{HEXA}  & 800   & 71,7  & - \\
    \textbf{RegionCL-M} & \textbf{800}   & \textbf{73.1}  & \textbf{91.5} \\
    \hline
    MoCov2*~\cite{chen2021exploring} & 800   & 72.3  & - \\
    \textbf{RegionCL-M*} & \textbf{800}   & \textbf{73.9}  & \textbf{92.0} \\
    \hline
    MoCov3~\cite{MoCov3} & 1000  & 74.6  & - \\
    \textbf{RegionCL-M3} & \textbf{1000}  & \textbf{75.4}  & \textbf{92.6} \\
    \hline
    \end{tabular}%
}
    }
    \vss
    \tablebox
    {
    \caption{Linear classification results on ImageNet~\cite{deng2009imagenet} using 1\% and 10\% data.}
    \label{tab:CLSFew}
    }
    {
  \setlength{\tabcolsep}{0.01\linewidth}{
  \begin{tabular}{l|ll|ll}
    \hline
          & \multicolumn{2}{c|}{1\% Data} & \multicolumn{2}{c}{10\% Data} \\
          & Top-1 & Top-5 & Top-1 & Top-5 \\
    \hline
    MoCov2~\cite{chen2020improved} & 43.6  & 70.9  & 58.8  & 82.4  \\
    \textbf{\modelName-M} & \textbf{46.1}  & \textbf{72.9}  & \textbf{60.4}  & \textbf{83.5}  \\
    \hline
    DenseCL~\cite{wang2021dense} & 38.9  & 66.2  & 54.0  & 79.3  \\
    \textbf{\modelName-D} & \textbf{47.8}  & \textbf{74.0}  & \textbf{60.4}  & \textbf{83.1}  \\
    \hline
    SimSiam~\cite{chen2021exploring} & 32.8  & 61.5  & 51.8  & 77.7  \\
    \textbf{\modelName-S} &  \textbf{42.3} & \textbf{70.6}  & \textbf{59.9} & \textbf{83.8}  \\
    \hline
    \end{tabular}
    }
    }
    }
    \end{floatrow}
\end{table}

\subsection{Detection and segmentation on MS COCO}
\noindent \textbf{Settings.} We show the detection performance of the models pretrained with the \modelName pretext task. The experiments are conducted on the MS COCO dataset~\cite{chen2015microsoft}, which contains about 118K images with bounding boxes and instance segmentation annotations and covers 80 object categories in total. We choose two representative detectors: the two-stage detector Mask-RCNN~\cite{he2017mask} and the one-stage detector RetinaNet~\cite{lin2017focal}, following the same settings as in \cite{xie2021detco,zhao2021self}. 

\begin{table}[htbp]
  \centering
  \caption{Object detection results on the MS COCO~\cite{chen2015microsoft} dataset with Mask-RCNN~\cite{he2017mask} C4 and FPN (1x).}\label{tab:COCO_C4_FPN_1X}
    \setlength{\tabcolsep}{0.002\linewidth}{\begin{tabular}{l|lll|lll|lll|lll}
    \hline
          & \multicolumn{6}{c|}{Mask-RCNN-C4-1x}           & \multicolumn{6}{c}{Mask-RCNN-FPN-1x} \\
          & AP$^{bb}$    & AP$_{50}^{bb}$  & AP$_{75}^{bb}$  & AP$^{mk}$    & AP$_{50}^{mk}$  & AP$_{75}^{mk}$  & AP$^{bb}$    & AP$_{50}^{bb}$  & AP$_{75}^{bb}$  & AP$^{mk}$    & AP$_{50}^{mk}$  & AP$_{75}^{mk}$ \\
    \hline
    Rand Init & 26.4  & 44.0  & 6.9   & 7.6   & 14.8  & 7.2   & 31.0  & 49.5  & 33.2  & 28.5  & 46.8  & 30.4  \\
    Supervised & 38.2  & 58.2  & 41.2  & 33.3  & 54.7  & 35.2  & 38.9  & 59.6  & 42.7  & 35.4  & 56.5  & 38.1  \\
    \hline
    InsDis~\cite{wu2018unsupervised} & 37.7  & 57.0  & 40.9  & 33.0  & 54.1  & 35.2  & 37.4  & 57.6  & 40.6  & 34.1  & 54.6  & 36.4  \\
    PIRL~\cite{misra2020self}  & 37.4  & 56.5  & 40.2  & 32.7  & 53.4  & 34.7  & 37.5  & 57.6  & 41.0  & 34.0  & 54.6  & 36.2  \\
    SwAV~\cite{caron2020unsupervised}  & 32.9  & 54.3  & 34.5  & 29.5  & 50.4  & 30.4  & 38.5  & 60.4  & 41.4  & 35.4  & 57.0  & 37.7  \\
    MoCo~\cite{he2020momentum}  & 38.5  & 58.3  & 41.6  & 33.6  & 54.8  & 35.6  & 38.5  & 58.9  & 42.0  & 35.1  & 55.9  & 37.7  \\
    DetCo~\cite{xie2021detco} & 39.4  & 59.2  & 42.3  & 34.4  & 55.7  & 36.6  & 39.5  & 60.3  & 43.1  & 35.9  & 56.9  & 38.6 \\
    DetCo-AA~\cite{xie2021detco} & 39.8  & 59.7  & 43.0  & 34.7  & 56.3  & 36.7  & 40.1  & 61.0  & 43.9  & 36.4  & 58.0  & 38.9 \\
    \hline
    MoCo v2~\cite{chen2020improved} & 38.9  & 58.4  & 42.0  & 34.2  & 55.2  & 36.5  & 38.9  & 59.4  & 42.4  & 35.5  & 56.5  & 38.1  \\
    \textbf{\modelName-M} & \textbf{39.8} & \textbf{59.8} & \textbf{43.0} & \textbf{34.8} & \textbf{56.4} & \textbf{36.9} & \textbf{40.1} & \textbf{60.7} & \textbf{43.9} & \textbf{36.3} & \textbf{57.7} & \textbf{39.0} \\
    \hline
    DenseCL~\cite{wang2021dense} & 39.3  & 59.1  & 42.2  & 34.5  & 55.6  & 36.8  & 39.1  & 59.4  & 42.5  & 35.5  & 56.4  & 38.0  \\
    \textbf{\modelName-D} & \textbf{40.3} & \textbf{60.3} & \textbf{43.9} & \textbf{35.2} & \textbf{57.0} & \textbf{37.3} & \textbf{40.4} & \textbf{61.3} & \textbf{44.2} & \textbf{36.7} & \textbf{58.2} & \textbf{39.4} \\
    \hline
    SimSiam~\cite{chen2021exploring} & 37.9  & 57.5  & 40.9  & 33.2  & 54.2  & 35.2  & 37.3  & 57.2  & 40.5  & 33.9  & 54.2  & 36.1  \\
    \textbf{\modelName-S} & \textbf{38.7} & \textbf{58.2} & \textbf{41.3} & \textbf{33.7} & \textbf{55.0} & \textbf{35.6} & \textbf{38.8} & \textbf{58.8} & \textbf{42.4} & \textbf{35.2} & \textbf{56.0} & \textbf{37.6} \\
    \hline
    \end{tabular}}%
\end{table}%

\noindent \textbf{Results of Mask-RCNN on MS COCO.}
Table~\ref{tab:COCO_C4_FPN_1X} and \ref{tab:COCO_C4_FPN_2X} summarize the Mask-RCNN results on 1x and 2x schedules respectively, where the \modelName variants are highlighted in bold. We can see that \modelName has significantly improved all approaches with ResNet50-C4 and ResNet50-FPN backbones, confirming
{the benefits of {region-level contrastive learning} on various SSL methods.}
According to the tables, incorporating \modelName into MoCov2 (\modelName-M) can further improve the performance over the MoCov2 baseline with both backbones. It is also noticeable that \modelName-M has already surpassed the previous representative SSL methods designed for dense prediction, \eg, DetCo~\cite{xie2021detco} and DenseCL~\cite{wang2021dense}. More importantly, when incorporating \modelName into DenseCL, \modelName-D achieves the best scores for all metrics in both the 1x and 2x settings. {It suggests that {RegionCL} can still help the dense prediction-favored methods to learn more discriminative features {from a complete perspective}.}

\begin{table}[htbp]
  \centering
  \caption{Object detection results on the MS COCO~\cite{chen2015microsoft} dataset with Mask-RCNN~\cite{he2017mask} C4 and FPN (2x).}\label{tab:COCO_C4_FPN_2X}
    \setlength{\tabcolsep}{0.002\linewidth}{\begin{tabular}{l|lll|lll|lll|lll}
    \hline
          & \multicolumn{6}{c|}{Mask-RCNN-C4-2x}           & \multicolumn{6}{c}{Mask-RCNN-FPN-2x} \\
         & AP$^{bb}$    & AP$_{50}^{bb}$  & AP$_{75}^{bb}$  & AP$^{mk}$    & AP$_{50}^{mk}$  & AP$_{75}^{mk}$  & AP$^{bb}$    & AP$_{50}^{bb}$  & AP$_{75}^{bb}$  & AP$^{mk}$    & AP$_{50}^{mk}$  & AP$_{75}^{mk}$ \\
    \hline
    Rand Init & 35.6  & 54.6  & 38.2  & 31.4  & 51.5  & 33.5  & 36.7  & 56.7  & 40.0  & 33.7  & 53.8  & 35.9  \\
    Supervised & 40.0  & 59.9  & 43.1  & 34.7  & 56.5  & 36.9  & 40.6  & 61.3  & 44.4  & 36.8  & 58.1  & 39.5  \\
    \hline
    MoCo~\cite{he2020momentum}  & 40.7  & 60.5  & 44.1  & 35.4  & 57.3  & 37.6  & 40.8  & 61.6  & 44.7  & 36.9  & 58.4  & 39.7  \\
    MaskCo~\cite{zhao2021self} & 40.8  & 60.5  & 44.2  & 35.5  & 57.1  & 38.0  & -     & -     & -     & -     & -     & - \\
    UnMix~\cite{UnMix} & - & - & - & - & - & - & 41.2 & 60.9 & 44.7 & - & - & - \\
    DetCo~\cite{xie2021detco} & 41.4  & 61.2  & 44.7  & 35.8  & 57.8  & 38.3  & 41.5  & 62.1  & 45.6  & 37.6  & 59.2  & 40.5  \\
    DetCo-AA~\cite{xie2021detco} & 41.3  & 61.2  & 45.0  & 35.8  & 57.9  & 38.2  & 41.5  & 62.5  & 45.6  & 37.7  & 59.5  & 40.5  \\
    \hline
    MoCo v2~\cite{chen2020improved} & 41.0  & 60.6  & 44.5  & 35.6  & 57.2  & 38.0  & 40.9  & 61.5  & 44.7  & 37.0  & 58.7  & 39.8  \\
    \textbf{\modelName-M} & \textbf{41.5} & \textbf{61.4} & \textbf{44.9} & \textbf{35.9} & \textbf{57.7} & \textbf{38.6} & \textbf{41.6} & \textbf{62.5} & \textbf{45.6} & \textbf{37.7} & \textbf{59.3} & \textbf{40.4} \\
    \hline
    DenseCL~\cite{wang2021dense} & 39.7 & 59.1 & 43.0 & 34.5 & 55.9 & 37.3 & 41.4  & 62.1  & 45.1  & 37.5  & 58.8  & 40.3  \\
    \textbf{\modelName-D} & \textbf{41.8} & \textbf{61.6} & \textbf{45.4} & \textbf{36.4} & \textbf{58.5}  & \textbf{39.2} & \textbf{42.1} & \textbf{62.9} & \textbf{45.9} & \textbf{38.0} & \textbf{60.0} & \textbf{40.7} \\
    \hline
    SimSiam~\cite{chen2021exploring} & 38.8  & 58.0  & 41.9  & 34.0  & 55.1  & 36.2  & 40.1  & 60.6  & 43.8  & 36.4  & 57.7  & 39.1  \\
    \textbf{\modelName-S} & \textbf{40.7} & \textbf{60.4} & \textbf{44.4} & \textbf{35.4} & \textbf{57.0} & \textbf{37.7} & \textbf{41.0} & \textbf{61.6} & \textbf{44.7} & \textbf{37.1} & \textbf{58.6} & \textbf{39.8} \\
    \hline
    \end{tabular}}%
  %
\end{table}%

\noindent \textbf{Results of RetinaNet on MS COCO.}
The results of RetinaNet on MS COCO using different SSL methods are presented in Table~\ref{tab:COCO_Retina}. From the table, we can see that the improvement brought by \modelName still holds in all metrics and at all the training settings. Similarly, \modelName-D achieves the best results at 38.8 AP and 40.6 AP for the two training schedules respectively, significantly surpassing the supervised baseline by 1.4 AP and 1.7 AP. It is also noted that the improvement in the more stringent metric AP$_{75}^{bb}$ is more significant than that in the AP$^{bb}$ metric, demonstrating that leveraging both cropped and remained regions for contrastive learning contributes to learning better feature representations for object detection and thus improving the detection accuracy. {These results show that the simple strategy \modelName can help existing SSL methods achieve a better trade-off between the classification and detection tasks (see Figure~\ref{fig:opening}), further validating the benefits of using both regions for pretraining}.

\subsection{Segmentation on Cityscapes}

\noindent \textbf{Settings.} Further, we evaluate the models' transfer performance for both instance and semantic segmentation on the Cityscapes~\cite{Cordts2016Cityscapes} dataset, which contains over 5K well-annotated images of street scenes from 50 different cities. We follow the same setting as in MoCov2~\cite{he2020momentum} for instance segmentation, with Mask-RCNN and trained for 24K iterations. UPerNet~\cite{xiao2018unified} in mmseg~\cite{mmseg2020} is employed for semantic segmentation evaluation.

\noindent \textbf{Results on Cityscapes.}
Table~\ref{tab:cityscape_upernet} presents the performance of different SSL methods and their variants with \modelName. The second and third columns show the performance for instance and semantic segmentation, respectively. According to the table, \modelName consistently improves the three representative SSL methods by large margins. For example, \modelName-M reaches the best on instance segmentation at 34.9 AP and 62.5 AP$_{75}$, while \modelName-D outperforms the others on semantic segmentation at 78.7 mIoU and 79.5 mIoU with different training schedules, {confirming the generalization and the effectiveness of {region-level contrastive learning with remained regions} on segmentation tasks.}

{
\subsection{Generalization on vision transformer}
To further evaluate the benefits of using both regions for pretraining, {we apply \modelName on vision transformer-based methods and train them following the design in MoBy~\cite{MOBY}.} {As shown in Table~\ref{tab:ViT}, RegionCL improves MoBy by over 3.0 accuracy on both ImageNet and ImageNet Real with 100 epochs pretraining and over 1.5 accuracy with 300 epochs pretraining.} It can be concluded that \modelName can successfully improve the vision transformer's performance, validating its effectiveness for advanced neural architectures like vision transformers.
}

\thisfloatsetup{heightadjust=all,valign=c}
\begin{table}[htbp]
\begin{floatrow}[2]
\tablebox{
  \caption{Detection results on MS COCO~\cite{chen2015microsoft} with RetinaNet~\cite{lin2017focal}.}
  \label{tab:COCO_Retina}
  }
  {%
  {
  \setlength{\tabcolsep}{0.00001\linewidth}{\begin{tabular}{l|ccc|ccc}
    \hline
          & \multicolumn{3}{c|}{RetinaNet-1x} & \multicolumn{3}{c}{RetinaNet-2x} \\
         & AP$^{bb}$    & AP$_{50}^{bb}$  & AP$_{75}^{bb}$  & AP$^{bb}$    & AP$_{50}^{bb}$  & AP$_{75}^{bb}$ \\
    \hline
    Rand Init & 24.5  & 39.0  & 25.7  & 32.2  & 49.4  & 34.2  \\
    Supervised & 37.4  & 56.5  & 39.7  & 38.9  & 58.5  & 41.5  \\
    \hline
    InsDis~\cite{wu2018unsupervised} & 35.5  & 54.1  & 38.2  & 38.0  & 57.4  & 40.5  \\
    PIRL~\cite{misra2020self}  & 35.7  & 54.2  & 38.4  & 38.5  & 57.6  & 41.2  \\
    SwAV~\cite{caron2020unsupervised}  & 35.2  & 54.9  & 37.5  & 38.6  & 58.8  & 41.1  \\
    MoCo~\cite{he2020momentum}  & 36.3  & 55.0  & 39.0  & 38.7  & 57.9  & 41.5  \\
    DUPR~\cite{ding2021unsupervised} & 38.0 & 57.2 & 40.7 & 40.0 & 59.6 & 43.0 \\
    DetCo~\cite{xie2021detco} & 38.0  & 57.4  & 40.7  & 39.8  & 59.5  & 42.4  \\
    DetCo-AA~\cite{xie2021detco} & 38.4  & 57.8  & 41.2  & 39.7  & 59.3  & 42.6  \\
    \hline
    MoCov2~\cite{chen2020improved} & 37.2  & 56.2  & 39.6  & 39.3  & 58.9  & 42.1  \\
    \textbf{\modelName-M} & \textbf{38.4}  & \textbf{58.1}  & \textbf{41.2}  & \textbf{40.1}  & \textbf{59.9}  & \textbf{43.2}  \\
    \hline
    DenseCL~\cite{wang2021dense} & 37.7  & 56.4  & 40.2  & 39.8  & 59.2  & 42.8  \\
    \textbf{\modelName-D} & \textbf{38.8}  & \textbf{58.6}  & \textbf{41.6}  & \textbf{40.6}  & \textbf{60.4}  & \textbf{43.6} \\
    \hline
    SimSiam~\cite{chen2021exploring} & 35.5  & 53.7  & 38.1  & 38.1 & 57.4 & 40.8 \\
    \textbf{\modelName-S} & \textbf{36.8} & \textbf{55.9} & \textbf{39.5} &  \textbf{39.1}  & \textbf{58.5} & \textbf{41.8} \\
    \hline
    \end{tabular}}}%
  }%
\tablebox{
    \caption{Instance (Inst.) and semantic (Sem.) segmentation results (mIoU) on Cityscapes~\cite{Cordts2016Cityscapes}.}
    \label{tab:cityscape_upernet}
    }
  {%
  {\setlength{\tabcolsep}{0.00001\linewidth}{\begin{tabular}{l|cc|cc}
    \hline
          & \multicolumn{2}{c|}{Inst. Seg} & \multicolumn{2}{c}{Sem. Seg} \\
    \cline{2-5}
          & \multicolumn{1}{c}{AP} & \multicolumn{1}{c|}{AP$_{75}$} & \makecell[c]{40K} & \makecell[c]{80K} \\
    \hline
    Supervised & 32.9  & 59.6  & 77.1  & 78.2  \\
    \hline
    MoCov2~\cite{chen2020improved} & 33.9  & 60.8  & 77.8  & 78.6  \\
    \textbf{\modelName-M} & \textbf{34.9} & \textbf{62.5} & \textbf{78.1} & \textbf{79.0} \\
    \hline
    DenseCL~\cite{wang2021dense} & 34.5  & 61.9  & 78.3  & 79.1  \\
    \textbf{\modelName-D} & \textbf{34.8}  & \textbf{62.3}  & \textbf{78.7} & \textbf{79.5} \\
    \hline
    SimSiam~\cite{chen2021exploring} & 33.6  & 61.0    & 76.2  & 78.1  \\
    \textbf{\modelName-S} & \textbf{34.9} & \textbf{61.6} & \textbf{77.8} & \textbf{78.7} \\
    \hline
    \end{tabular}}}%
  }
\end{floatrow}
\end{table}

\subsection{Ablation Study}

We conduct the ablation studies with \modelName-M. All models are trained for 100 epochs {due to the limitation of computation resources and follow the practice of previous works~\cite{xie2021detco,zhao2021self}.} We adopt a $k$-NN classifier to evaluate their classification accuracy on ImageNet~\cite{deng2009imagenet} and train these models for 12K iterations on MS COCO~\cite{chen2015microsoft} to evaluate their dense prediction performance.

\begin{table}
\floatbox[{\capbeside\thisfloatsetup{capbesideposition={right,center},capbesidewidth=0.4\textwidth}}]{table}[\FBwidth]
{\caption{Linear classification results comparsion on ImageNet with vision transformer-based method MoBy~\cite{MOBY}.}\label{tab:ViT}}
{
    \setlength{\tabcolsep}{0.00001\linewidth}{\begin{tabular}{l|cc|cc|c}
    \hline
          & \multirow{2}[1]{*}{Backbone} & \multirow{2}[1]{*}{Epochs} & \multicolumn{2}{c|}{ImageNet} & \multicolumn{1}{c}{Real} \\
          & & & Top-1 & \multicolumn{1}{c|}{Top-5} & \multicolumn{1}{c}{Top-1} \\
    \hline
    MoBy~\cite{MOBY}  & Swin-T & 100   & 70.9  &  89.7  & 77.5 \\
    RegionCL+MoBy & Swin-T & 100   & 73.9  & 91.8 & 81.2 \\
    MoBy~\cite{MOBY}  & Swin-T & 300   & 75.3  & 92.2 & 82.4 \\
    RegionCL+MoBy & Swin-T & 300   & 77.0  & 93.1 & 83.9  \\
    \hline
    \end{tabular}}%
}
\end{table}

\thisfloatsetup{heightadjust=all,valign=t}
\begin{table}[htbp]
\begin{floatrow}[2]
\tablebox{
  \caption{The influence of paste view's size. $\mathcal{C}_L$ and $\mathcal{C}_U$ denote the lower and upper bound of the size for the paste view generation. `-' denotes no paste view is used during the training, \ie, downgrading to the original MoCov2.}
  \label{tab:ablation_range}
  }
  {%
  {
  \setlength{\tabcolsep}{0.00001\linewidth}{\begin{tabular}{cc|cc|cc}
    \hline
    \multicolumn{2}{c|}{Configuration} & \multicolumn{2}{c|}{ImageNet} & \multicolumn{2}{c}{MS COCO} \\
    \hline
    $\mathcal{C}_L$ & $\mathcal{C}_U$ & 20-NN & 100-NN & AP$^{bb}$ & AP$^{mk}$ \\
    \hline
    -     & -     & 49.3  & 47.3  & 26.3  & 24.0  \\
    \hline
    1     & 5     & 51.8  & 49.8  & 27.3  & 24.9  \\
    2     & 5     & 53.0  & 51.2  & 27.9  & 25.5  \\
    \textbf{3}     & \textbf{5}     & \textbf{54.7}  & \textbf{52.7}  & \textbf{28.0}  & \textbf{25.6}  \\
    4     & 5     & 53.9  & 51.8  & 27.9  & 25.5  \\
    \hline
    3     & 4     & \textbf{55.1}  & \textbf{52.8}  & 27.8  & 25.5  \\
    3     & 6     & 54.3  & 52.5  & 27.8  & 25.5  \\
    \hline
    \end{tabular}}}%
  }%
\tablebox{
    \caption{The influence of paste and canvas views. `Paste'/`Canvas' denote using paste/canvas views as positive pairs. `Neg' means using the canvas and paste counterpart views in the composited images as negative pairs.}
    \label{tab:ablation_views}
    }
  {%
  \setlength{\tabcolsep}{0.001\linewidth}{\begin{tabular}{ccc|cc|cc}
    \hline
    \multicolumn{3}{c|}{Configuration} & \multicolumn{2}{c|}{ImageNet} & \multicolumn{2}{c}{MS COCO} \\
    \hline
    Paste & Canvas & Neg & 20-NN & 100-NN & AP$^{bb}$ & AP$^{mk}$ \\
    \hline
     $\times$   & $\times$  & $\times$  &   49.3    &   47.3    &   26.3    &   24.0 \\
     $\times$    & \checkmark     & $\times$  & 51.8  & 50.0  & 26.9  & 24.7 \\   
    \checkmark     &   $\times$    & $\times$     & 50.0  & 50.2  & 27.0  & 24.7  \\
    \checkmark     & \checkmark     &   $\times$    & 54.6  & 52.4  & 27.8  & 25.5  \\
    \hline
    \checkmark     & \checkmark     & \checkmark     & 54.7  & 52.7  & 28.0  & 25.6  \\
    \hline
    \end{tabular}}%
  }
\end{floatrow}
\end{table}

\noindent \textbf{The size of the paste view.} We investigate the influence of the size of the paste view by varying the lower and upper bounds $\mathcal{C}_L$ and $\mathcal{C}_U$. Note that the image size is set as 224 during the training and the downsampling ratio for the backbone network ResNet-50~\cite{he2016deep} is 32, thus $\mathcal{C}_L,\mathcal{C}_U$ are valid in the range $[1, 7]$. The optimal hyper-parameters are determined through two steps.
\textbf{(1)} We first fix the upper bound $\mathcal{C}_U$ to 5 and search different configurations for the lower bound $\mathcal{C}_L$. As shown in Table~\ref{tab:ablation_range}, the performance on both classification and detection peaks with $\mathcal{C}_L=3$.
\textbf{(2)} Then we fix $\mathcal{C}_L$ to 3 and search for $\mathcal{C}_U$. It is interesting to see that decreasing $\mathcal{C}_U$ from 5 to 4 slightly improves the performance on classification but degrades that on detection. This suggests that the optimal configurations of $\mathcal{C}_L,\mathcal{C}_U$ for classification and dense prediction tasks may be slightly different, and we select $\mathcal{C}_L=3, \mathcal{C}_U=5$ as default values to achieve a trade-off on both classification and dense prediction tasks. 

\noindent \textbf{The influence of using paste and canvas views.} We further investigate the importance of using both paste and canvas views during pretraining. The results are concluded in Table~\ref{tab:ablation_views}, where \checkmark under Paste or Canvas denotes whether to use the former or latter term in Eq.~\eqref{eq:Lp}. The `Negative' option means whether to treat the canvas and paste counterpart views from the same composite image $x^{pc}$ as negative pairs, \ie, $\exp{(c \cdot sg(p) / \tau} )$ or $\exp{(p \cdot sg(c) / \tau} )$ in the denominator in Eq.~\eqref{eq:Lp}. With all columns marked $\times$, the method becomes standard MoCov2. From the first three rows in the table, we can see that using either the paste or canvas views can bring performance gains, and the performance will be further boosted when considering both regions. For example, in the 4th row, the model gains more than 5\% accuracy improvement over MoCov2 in both ImageNet 20- and 100-NN classification. We attribute it to that leveraging both regions during pretraining can help models learn better category features from a complete perspective. Comparing the 4th row with the last row, where intra-image negative pairs are utilized, the performance on both tasks is slightly improved. It demonstrates that using negative pairs within images can help models learn more discriminative features between different regions, again validating the importance of introducing region-level contrastive pairs in self-supervised learning.

\subsection{Analysis of \modelName}

{
\begin{figure}
\floatbox[{\capbeside\thisfloatsetup{capbesideposition={right,center},capbesidewidth=0.4\textwidth}}]{figure}[\FBwidth]
{\caption{The KNN accuracy and average gradient magnitude of MoCov2~\cite{chen2020improved} and \modelName-M with different training epochs.}\label{fig:trainCurve}}
{\includegraphics[width=0.47\textwidth]{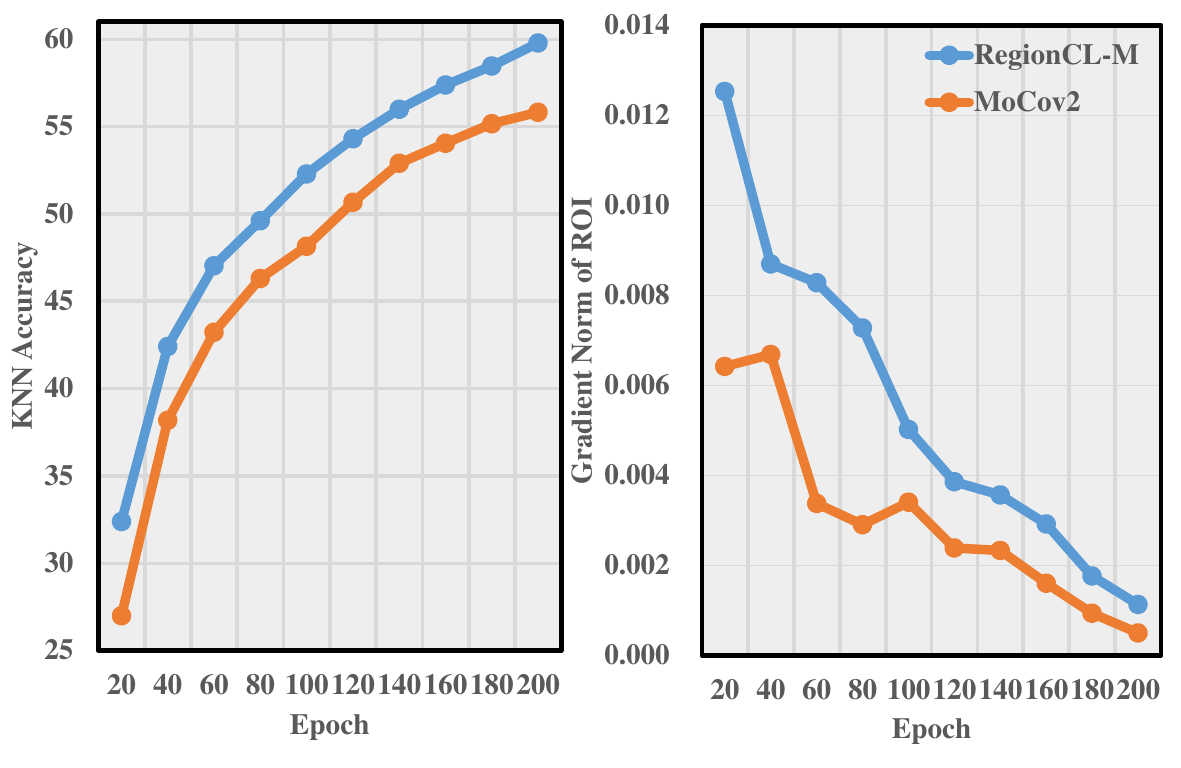}}
\end{figure}
To better analyze the performance gains brought by \modelName, we monitor the KNN accuracy and gradients of the target regions during training. We use YOLOX~\cite{yolox2021} to detect the objects in the images and record the average magnitude of gradients back-propagated from the loss of MoCov2 and \modelName-M in the object bounding boxes on the training set. As shown in Figure~\ref{fig:trainCurve}, the KNN accuracy of \modelName-M consistently outperforms MoCov2 while the gradients inside the object regions of \modelName-M are larger than those of MoCov2, implying that using both regions for training can help models pay more attention on the targets and learn better object representation from a complete perspective.
}

\section{Limitation and Discussion}
{
We make the first attempt to demonstrate the importance of considering both cropped and remained regions in SSL via a simple yet effective \modelName pretext task. The simple task makes minor modifications to representative SSL methods to show the benefits brought by leveraging regional contrastive pairs. Although \modelName effectively improve these methods, there is still much to be explored in utilizing the remained regions for more tasks~\cite{xu2022dut,xu2022vitpose,li2022exploring,bao2021beit}, \eg, introducing additional bounding box selection and alignment modules or adopting multi-level supervision for learning. Besides, RegionCL temporally costs more computations per iteration (about 50\% for RegionCL-M). Although it demonstrates better results with fewer computation resources as discussed in the supplementary, it will be beneficial to reduce the training costs, which we leave as our future work.
}

\section{Conclusion}
This paper demonstrates the importance of using both cropped and remained regions after cropping for self-supervised learning. {A simple yet effective pretext task \modelName is proposed to validate the models can learn better category feature representation from a complete perspective.} Experimental results on image classification, object detection, and instance and semantic segmentation benchmarks demonstrate the effectiveness of leveraging remained regions in pretraining and its compatibility to representative self-supervised learning methods. We hope that this study will provide valuable insights into the subsequent studies of self-supervised learning in exploring region-based contrast methodology.

\noindent \textbf{Acknowledgement} Mr. Yufei Xu, Mr. Qiming Zhang, and Dr. Jing Zhang are supported by ARC FL-170100117.

\appendix

\section{Appendix}

{
In the appendix, we demonstrate the benefits of leveraging both cropped and remained regions for training with longer training epochs (\ref{subsec:epochs}) and larger models (\ref{subsec:sizes}). We also demonstrate the training efficiency of RegionCL in section~\ref{subsec:epochs}. Pose estimation for both human and animals are also employed to evaluate the effectiveness of region-level contrastive learning on downstream tasks (\ref{subsec:pose}). Besides, we also independently feed the paste and canvas views into the networks for training to further validate the performance of regional contrastive pairs without swapping (\ref{subsec:noswapping}). The analysis of parameters, architecture details, and implementation details are provided in \ref{sec:bz}, \ref{subsec:architecture}, and \ref{subsec:implementation}, respectively. 
}

\subsection{Training efficiency with longer training epochs}\label{subsec:epochs}

\begin{table}[htbp]
  \centering
  \footnotesize
  \caption{Results of MoCo v2~\cite{chen2020improved} and \modelName-M trained for 200, 400, and 800 epochs. `*' denotes that we end-to-end finetune \modelName-M pretrained models for 50 epochs~\cite{he2021masked,cai2021exponential}.}
    \setlength{\tabcolsep}{0.012\linewidth}{\begin{tabular}{l|lll|lll|lll}
    \hline
          & \multicolumn{3}{c|}{ImageNet Top-1} & \multicolumn{3}{c|}{MS COCO AP$^{bb}$} & \multicolumn{3}{c}{MS COCO AP$^{mk}$ } \\
    Epochs  & 200   & 400   & 800   & 200   & 400   & 800   & 200   & 400   & 800 \\
    \hline
    MoCo v2 & 67.5  &  69.6  & 71.1  & 40.9  & 41.3  & 41.5  & 37.0  & 37.5  & 37.6 \\
    \modelName-M & 69.4  & 72.1  & 73.1  & 41.6  & 41.9  & 42.1  & 37.7  & 38.0  & 38.2  \\
    \modelName-M* & 76.8  & 77.8  & 78.1  & -  & -  & -  & -  & -  & -  \\
    \hline
    \end{tabular}}%
  \label{tab:supp_longer}%
\end{table}%

{We investigate effectiveness of leveraging remained regions for longer pretraining by extending the epochs to 200, 400, and 800 respectively.} We train MoCov2~\cite{chen2020improved} and the corresponding \modelName-M respectively and present their results in Table~\ref{tab:supp_longer}. We evaluate these models' image classification performance on the ImageNet~\cite{deng2009imagenet} dataset with linear probing. Their object detection and instance segmentation performance are also evaluated on the MS COCO~\cite{chen2015microsoft} dataset with ResNet50-FPN and Mask-RCNN~\cite{he2017mask}. The detection and segmentation models are trained following the 2$\times$ schedule, \ie, the models are trained for 180K iterations in total. 

As can be seen, with only 200 epochs for training, the proposed \modelName-M obtains competitive results compared with MoCov2 trained for 400 epochs, no matter on classification or dense prediction tasks. 
{The performance of \modelName-M increases with the total training epochs increasing, and \modelName-M can obtain better performance with less training cost, \ie, \modelName-M with 400 epochs pretraining has significantly outperformed MoCov2 trained with 800 epochs, confirming the good property of training efficiency brought by simply training with the regional contrastive pairs.} Although RegionCL costs more forwards each iterations than the baseline method, it improves the overall training efficiency, \ie, RegionCL with 400 epochs training (1200 forwards) beats the baseline methods with 800 epochs training (1600 forwards).

Besides, \modelName-M sees an further improvement especially for classification (by 1\% accuracy) when extending to 800 training epochs, reaching 73.1\% Top-1 accuracy for classification, 42.1 AP for object detection and 38.2 AP for instance segmentation. Such observation demonstrates that the abundant contrastive pairs with both cropped and remained regions can not only improves the model's convergence but also effectively enhance the model's representation capacity with more training epochs.

\subsection{Generalization ability for models with variant sizes}\label{subsec:sizes}

\begin{table}[htbp]
  \centering
  \footnotesize
  \caption{Results of MoCo v2~\cite{chen2020improved} and \modelName-M with R50~\cite{he2016deep}, R50w2~\cite{he2016deep}, and R50w4~\cite{he2016deep} as backbones. `*' denotes that we end-to-end finetune \modelName-M pretrained models for 50 epochs~\cite{he2021masked,cai2021exponential}.}
    \setlength{\tabcolsep}{0.012\linewidth}{\begin{tabular}{l|lll|lll|lll}
    \hline
          & \multicolumn{3}{c|}{ImageNet Top-1} & \multicolumn{3}{c|}{MS COCO AP$^{bb}$} & \multicolumn{3}{c}{MS COCO AP$^{mk}$} \\
    Models      & R50   & w2 & w4 & R50   & w2 & w4 & R50   & w2 & w4 \\
    \hline
    MoCo v2 & 67.5  & 72.0 & 74.1 & 40.9  &  43.2  & 43.9 & 37.0  & 38.7 &  39.4 \\
    RegionCL-M & 69.4  & 75.2 & 76.2  & 41.6  & 43.8 & 44.5 & 37.7  &  39.3 & 39.7 \\
    RegionCL-M* & 76.8  & 79.7 & 80.4  & -  & - & - & - & - & - \\
    \hline
    \end{tabular}}%
  \label{tab:supp_modelsize}%
\end{table}%

{To investigate the benefits of regional contrastive pairs on models with variant sizes}, we adopt ResNet50~\cite{he2016deep}, ResNet50-w2 (2$\times$parameters), and ResNet50-w4 (4$\times$parameters) as backbone networks and train them for 200 epochs with MoCov2 and \modelName-M, respectively. The results of linear probing on ImageNet along with object detection and instance segmentation on MS COCO are reported in Table~\ref{tab:supp_modelsize}. We use Mask-RCNN with ResNet50-FPN as the object detection and instance segmentation framework and train them for 180K iterations, following the 2$\times$ schedule. It can be observed that \modelName-M with ResNet50-w2 outperforms MoCov2 with ResNet50-w4 on image classification and obtains competitive performance on both object detection and instance segmentation tasks on MS COCO. \modelName-M with ResNet50-w4 obtains the best performance on all tasks. It indicates that the proposed \modelName method is scalable to large models and can improve their performance on both classification and dense prediction tasks, further validating the importance of using both cropped and left regions in self-supervised learning.

\subsection{Results on pose estimation}\label{subsec:pose}
\begin{table}[htbp]
  \centering
  \footnotesize
  \caption{Results of \modelName compatible models on human (MS COCO~\cite{chen2015microsoft}) and animal (AP-10K~\cite{yu2021ap}) pose estimation.}
    \begin{tabular}{l|cc}
    \hline
          & \multicolumn{1}{l}{MS COCO~\cite{chen2015microsoft}} & \multicolumn{1}{l}{AP-10K~\cite{yu2021ap}} \\
          & \multicolumn{1}{c}{AP} & \multicolumn{1}{c}{AP} \\
    \hline
    Supervised & 71.8  & 69.9  \\
    \hline
    MoCo v2~\cite{chen2020improved} & 72.0  & 70.1  \\
    RegionCL-M & 72.3  & 70.6  \\
    \hline
    DenseCL~\cite{wang2021dense} & 72.4  & 71.1  \\
    RegionCL-D & 72.6  & 72.1  \\
    \hline
    SimSiam~\cite{chen2021exploring} & 71.9  & 70.5  \\
    RegionCL-S & 72.2  & 71.6  \\
    \hline
    \end{tabular}%
  \label{tab:supp_pose}%
\end{table}%
Besides the evaluation on detection and segmentation tasks, we also evaluate the models' performance on both human pose estimation and animal pose estimation tasks on MS COCO~\cite{chen2015microsoft} and AP-10K~\cite{yu2021ap} datasets. We adopt SimpleBaseline~\cite{xiao2018simple} as the base pose estimation framework and utilizes backbone models pretrained by MoCov2~\cite{chen2020improved}, DenseCL~\cite{wang2021dense}, SimSiam~\cite{chen2021exploring}, and their \modelName compatible counterparts. We train these models for 210 epochs. We adopt an Adam optimizer with initial learning rate at 1e-4, which decreases by a factor of 10 at the 170 and 200 epochs respectively, following the same setting as in mmpose~\cite{mmpose2020}. The results are available in Table~\ref{tab:supp_pose}. It can be observed that the SSL pretrained models outperforms the supervised counterpart. Besides, with both cropped and left regions taken into consideration, \modelName improves the pretrained models' transfer performance on both pose estimation tasks, especially on the smaller animal pose dataset AP-10k. Such observation further demonstrates that exploiting supervisory signals from both instance and region levels can help the model obtain a better trade-off on both classification and dense prediction tasks.

\subsection{Influence of the region swapping operation}\label{subsec:noswapping}
\begin{table}[htbp]
  \centering
  \footnotesize
  \caption{Influence of the region swapping strategy.}
    \begin{tabular}{l|c|c}
    \hline
    \multicolumn{1}{c|}{Configuration} & \multicolumn{1}{c|}{ImageNet Top-1} & \multicolumn{1}{c}{MS COCO AP$^{bb}$} \\
    \hline
    MoCov2 & 67.5 & 40.9\\
    w/o swapping & 69.2  & 41.4  \\
    w/ swapping & 69.4  & 41.6 \\
    \hline
    \end{tabular}%
  \label{tab:supp_off}%
\end{table}%

{To further understand the performance gains brought by the abundant contrastive pairs, we adopt a simple \modelName variant without the swapping operation, \ie, simply cropping a region from candidate images as paste views and filling zeros into the cropped regions to formulate the canvas views. Using \modelName-M as the base, we train \modelName-M and its variant without the region swapping operation for 200 epochs, and evaluate their performance on the ImageNet~\cite{deng2009imagenet} dataset and on the MS COCO~\cite{chen2015microsoft} dataset with ResNet50-FPN and Mask-RCNN~\cite{he2017mask} following the 2$\times$ schedule}. 
The results are available in Table~\ref{tab:supp_off}. Without the region swapping operations, the proposed \modelName still improves the model's performance on both classification and dense prediction tasks, while the region swapping operation can further improve the performance on both tasks, \eg, 0.2\% accuracy gains for ImageNet Top-1 accuracy and 0.2 mAP gains for object detection. 
{It indicates that although taking both regions into consideration without swapping can facilitate the models learning, composing the hard negative samples, \ie, the paste and canvas views via swapping, in the same images can further help the model learn better and discriminative feature representations from both instance- and region-level pairs, since features of these two kinds of views share some context from each other during network forward calculation.}

\subsection{Influence of different batch size}
\label{sec:bz}
\begin{table}[htbp]
  \centering
  \footnotesize
  \caption{The influence of batch size of MoCo v2 and \modelName.}
    \setlength{\tabcolsep}{0.02\linewidth}{\begin{tabular}{c|cc|cc|cc}
    \hline
          &       &       & \multicolumn{2}{c|}{ImageNet} & \multicolumn{2}{c}{MS COCO} \\
          & \multicolumn{1}{c}{Batch Size} & \multicolumn{1}{c|}{LR} & \multicolumn{1}{c}{Top-1} & \multicolumn{1}{c|}{Top-5} & \multicolumn{1}{c}{AP$^{bb}$} & \multicolumn{1}{c}{AP$^{mk}$} \\
    \hline
    MoCo v2 & 256   & 0.03  & 67.5  & \multicolumn{1}{c|}{-} & 40.9  & 37.0  \\
    MoCo v2 & 1024  & 0.15  & 67.5  & 88.2  & 41.0  & 37.2  \\
    \hline
    RegionCL-M & 256   & 0.03  &  70.0 & 90.0 & 41.6  & 37.8  \\
    RegionCL-M & 1024  & 0.15  & 69.4  & 89.6  & 41.6  & 37.7  \\
    \hline
    \end{tabular}}%
  \label{tab:supp_bz}%
\end{table}%

As the number of negative pairs plays an important role in the InfoNCE~\cite{oord2018representation} loss and affects the pretrained model's performance as pointed in \cite{chen2020a}, MoCov2 maintains a huge memory queue to provide enough negative samples, which makes the calculation of the InfoNCE loss and the training process not coupled with the batch size. Thus we accelerate the training of MoCov2~\cite{chen2020improved} and \modelName-M by increasing the batch size from 256 to 1,024. We train the models for 200 epochs with an initial learning rate 0.15 (around linear growth w.r.t. the batch size) and a cosine learning rate scheduler, while the origin training setting is 200 epochs with an initial learning rate 0.03 and a cosine learning rate scheduler. The other settings are exactly the same, including the data augmentation strategies, optimizers, and the values of hyper-parameters. We validate the performance difference of the two training settings and present the results in Table~\ref{tab:supp_bz}. It can be observed that MoCov2's performance are consistent for both classification and dense prediction tasks with both training settings, confirming the rationality of the batch size to be 1,024 for MoCo~v2. Thus, we choose such batch size for MoCo~v2-based models in our paper. 

We also conduct similar experiments for \modelName-M as shown in the last two rows in Table~\ref{tab:supp_bz}. Similar conclusion can be observed in the evaluation of \modelName-M with different batch size for training. Besides, as we adopt the batch-wise implementation for region swapping, \modelName-M with small batch size and thus more iterations can see more diverse paste views and canvas views in terms of different sizes and locations, thus learning better feature representations. As a result, \modelName-M with a batch size of 256 obtains slightly better performance for image classification by 0.6\% Top-1 accuracy. Nevertheless, we choose the batch size of 1024 in this paper for acceleration purpose.

\subsection{Architecture details}\label{subsec:architecture}

We present the details of the proposed \modelName-M (MoCov2), \modelName-D (DenseCL), and \modelName-S (SimSiam) in this section. We also provide the pseudo codes for \modelName-M, \modelName-D, and \modelName-S as in Algorithm~\ref{alg:Code-M}, \ref{alg:code-D}, and \ref{alg:code-S}, respectively, with {\textbf{\textit{\color{red}{red}}}} color denoting the modifications of \modelName compared with the base architecture. 

\begin{figure*}[t!]
    \centering
    \includegraphics[width=1.\linewidth]{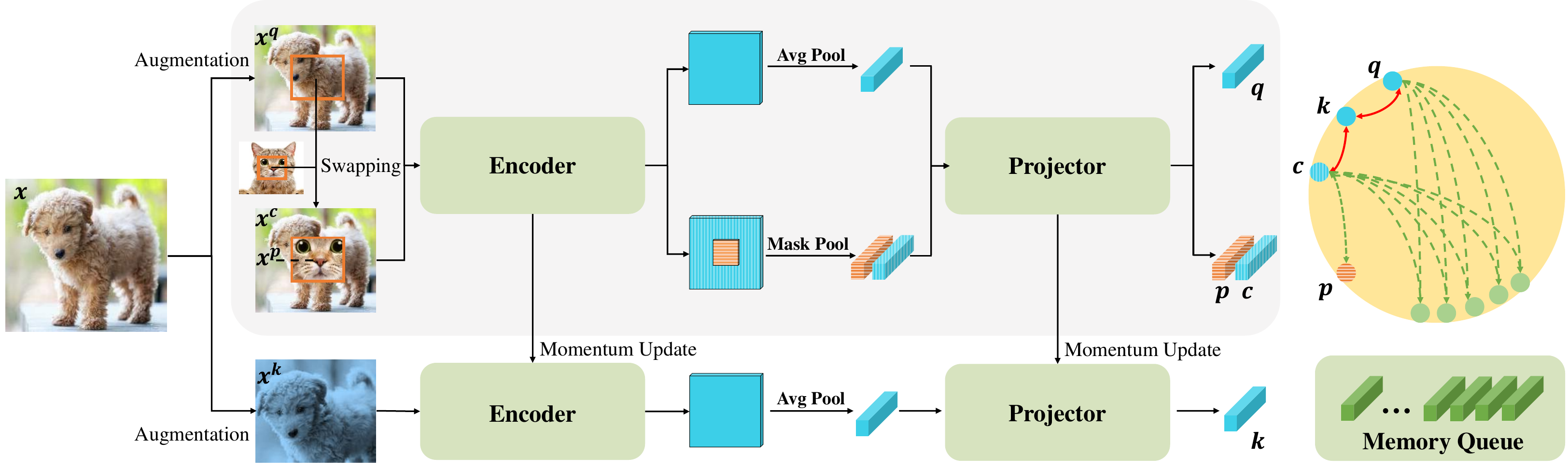}
    \vspace{-4 mm}
    \caption{Illustration of the proposed \modelName with the MoCov2 framework, \ie, \textbf{\modelName-M}. Taking the two augmented views $x^q$, $x^k$ as inputs, \modelName employs region swapping among the batch of $x^q$ to generate the composite images with paste views $x^p$ and canvas views $x^c$. Then, for the composite images, mask pooling is used to extract the features belonging to the paste and canvas views, respectively. The pooled region-level features (with stripes in the figure) are batched with the instance-level features and processed by the projector. The projected features $q$, $p$, $c$, $k$, and features from the memory queue form both instance- and region-level contrastive pairs.}
    \vspace{-2 mm}
    \label{fig:pipeline}
\end{figure*}

\begin{figure*}[t!]
    \centering
    \includegraphics[width=1.\linewidth]{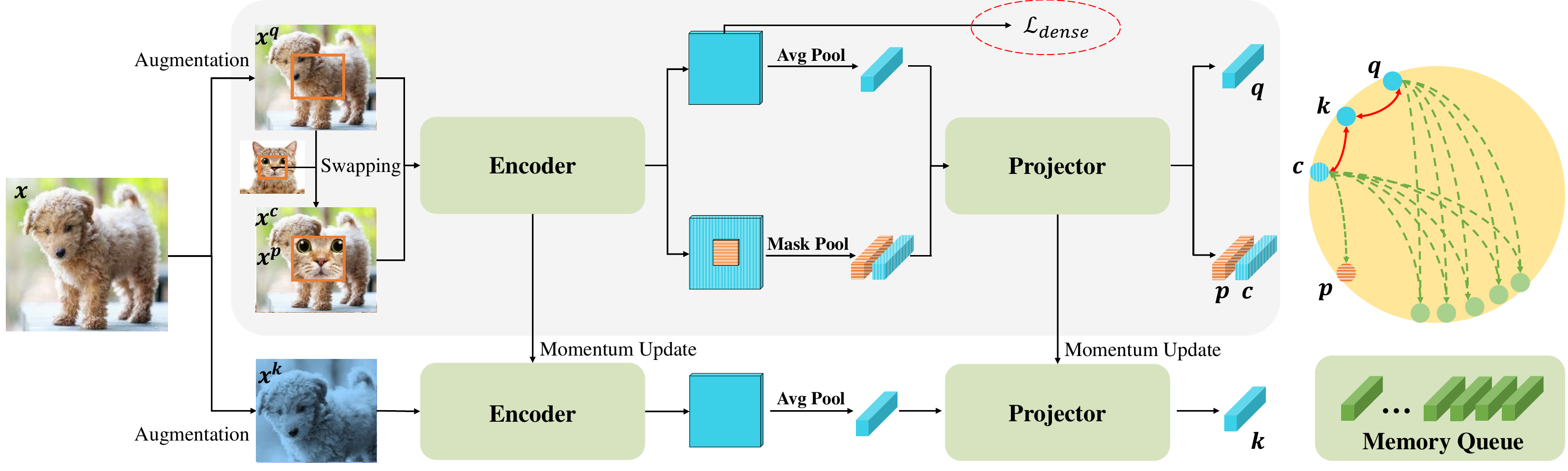}
    \vspace{-4 mm}
    \caption{Illustration of the proposed \modelName with the DenseCL framework, \ie, \textbf{\modelName-D}. Taken the instance-level views $x^q$ and $x^k$, and the region-level views $x^c$ and $x^p$ as inputs, \modelName-D firstly extract the instance- and region-level features using the same way as \modelName-M. The extracted and projected features $q$, $p$, $c$, and $k$ are constructed the contrastive pairs. The instance-level views $x^q$ and $x^k$ are also used to enhance dense feature correspondences before the average pooling operation, with another projector for pixel-wise feature projection and memory queue.}
    \vspace{-2 mm}
    \label{fig:supp_regionCLD}
\end{figure*}

\begin{figure*}[t!]
    \centering
    \includegraphics[width=1.\linewidth]{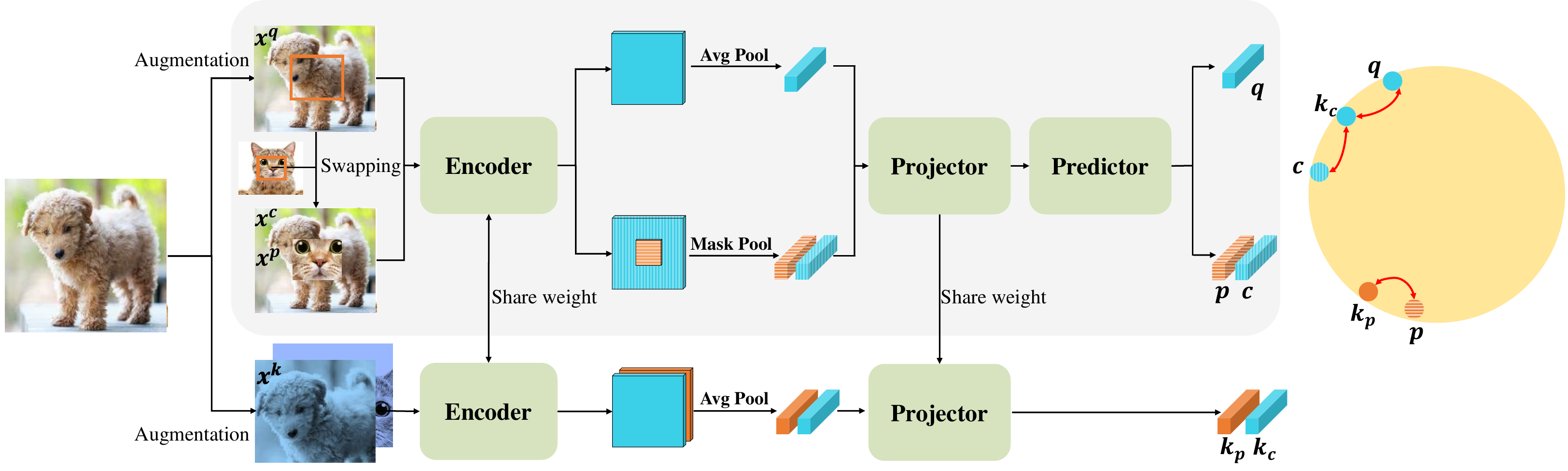}
    \vspace{-4 mm}
    \caption{Illustration of the proposed \modelName with the SimSiam framework, \ie, \textbf{\modelName-S}. Taking the two augmented views as inputs, \modelName-S extracts the region-level features in the same way as \modelName-M. The pooled region-level features are then batched with the instance-level features and processed by the projector. Following SimSiam, the projected features $q$, $p$, $c$, $k$ build positive pairs.}
    \vspace{-2 mm}
    \label{fig:supp_regionCLS}
\end{figure*}

\begin{algorithm}[htbp]
\small
\DontPrintSemicolon
  \KwInput{Two augmented views $x^q$, $x^k$}
  \KwInput{The negative Queue $Q$}
  \KwOutput{The contrastive loss $L$}
  
  \tcc{Feature extraction}
  
  \tcp{Online Branch}
  \difffont{$x^p_c$ = RegionSwapping($x^q$)}

  $q$ = Projector(AvgPool(Encoder($x^q$)))

  \difffont{$p$,$c$ = Projector(MaskPool(Encoder($x^p_c$)))}
  
  \tcp{Momentum Branch}
  $k$ = Projector$_M$(AvgPool(Encoder$_M$($x^k$)))
  
  \tcc{Loss computation}
  
  \tcp{Eq.~1}
  $L_{ins} = Loss(q, k | Q)$ 
  
  \tcp{Eq.~2}
  \difffont{$L_{reg} = (Loss(p, k|Q, sg(c)) + Loss(c, k|Q, sg(p))) / 2$} 
  
  $L$ = $L_{ins}$ + $L_{reg}$

\caption{Example code of \modelName-M.}
\label{alg:Code-M}
\end{algorithm}

\begin{algorithm}[htbp]
\small
\DontPrintSemicolon
  \KwInput{Two augmented views $x^q$, $x^k$}
  \KwInput{The instance negative Queue $Q_{ins}$}
  \KwInput{The dense negative Queue $Q_{dense}$}
  \KwOutput{The contrastive loss $L$}
  
  \tcc{Feature extraction}
  
  \tcp{Online Branch}
  \difffont{$x^p_c$ = RegionSwapping($x^q$)}

  $q$ = Projector(AvgPool(Encoder($x^q$)))

  \difffont{$p$,$c$ = Projector(MaskPool(Encoder($x^p_c$)))}
  
  \tcp{Extract dense feature}
  $q_d$ = Projector$_d$(Encoder($x^q$))

  \tcp{Momentum Branch}
  $k$ = Projector$_M$(AvgPool(Encoder$_M$($x^k$)))
  
  $k_d$ = Projector$_{dM}$(Encoder$_M$($x^k$))
  
  \tcc{Loss computation}
  
  \tcp{Eq.~1}
  $L_{ins} = Loss(q, k | Q_{ins})$ 
  
  $L_{dense} = DenseLoss(q_d, k_d | Q_{dense})$
  
  \tcp{Eq.~2}
  \difffont{$L_{reg} = (Loss(p, k|Q, sg(c)) + Loss(c, k|Q, sg(p))) / 2$} 
  
  $L$ = $L_{ins}$ + $L_{reg}$ + $L_{dense}$

\caption{Example code of \modelName-D.}
\label{alg:code-D}
\end{algorithm}

\begin{algorithm}[htbp]
\small
\DontPrintSemicolon
  \KwInput{Two augmented views $x^q$, $x^k$}
  \KwOutput{The contrastive loss $L$}
  
  \tcc{Feature extraction}
  
  \tcp{1st Branch}
  \difffont{$x^p_c$ = RegionSwapping($x^q$)}

  $q$ = Predictor(Projector(AvgPool(Encoder($x^q$))))

  \difffont{$p$,$c$ = Predictor(Projector(MaskPool(Encoder($x^p_c$))))}
  
  \tcp{2nd Branch, weight sharing with the 1st Branch}
  $k$ = Projector(AvgPool(Encoder($x^k$)))

  \tcc{Loss computation}
  
  \tcp{Eq.~1}
  $L_{ins} = Loss(q, sg(k))$ 

  \tcp{Eq.~2}
  \difffont{$L_{reg} = (Loss(p, sg(k)) + Loss(c, sg(k)) / 2$} 
  
  $L$ = $L_{ins}$ + $L_{reg}$

\caption{Example code of \modelName-S.}
\label{alg:code-S}
\end{algorithm}

\noindent \textbf{\modelName-D.} As shown in Algorithm~\ref{alg:code-D} and Figure~\ref{fig:supp_regionCLD}, DenseCL~\cite{wang2021dense} adopts both instance-level and pixel-level losses during pretraining. The modifications from DenseCL to \modelName-D appears at the instance-level branch in a same way as the modifications from MoCov2~\cite{chen2020improved} to \modelName-M, \ie, we use the encoder, mask pooling, and the instance-level projector to extract the features belonging to the canvas and paste views, separately. Then, the contrastive pairs are also enriched by the regions while the dense correspondences related loss functions are remained the same as in DenseCL. 

\noindent \textbf{\modelName-S.} As shown in Algorithm~\ref{alg:code-S} and Figure~\ref{fig:supp_regionCLS}, SimSiam~\cite{chen2021exploring} does not require the negative pairs during pretraining and focuses on attracting the features among positive pairs. The modifications from SimSiam~\cite{chen2021exploring} to \modelName-S are simply providing abundant positive pairs from both instance and region levels. Specifically, given the augmented views $x^q$, $x^k$ and the composite images with the canvas view $x^c$ and the paste view $x^p$ as inputs, \modelName-S adopts an encoder, average pooling (mask pooling) layer, a projector, and a predictor to get the instance-level (region-level) features $q$ ($p$ and $c$). The other view $x^k$ are processed by the weight-shared encoder and projector to get the feature $k$, where an stop gradient operation is applied on $k$ to stabilize the training. Thus, there are three cases of positive pairs in the modified \modelName-S, \ie, the instance-level pairs $q$ and $k$ as in origin SimSiam, the region-level pairs $c$ and $k_c$, and $p$ and $k_p$, and we keep the learning objectives and architecture the same as in SimSiam.

\subsection{Implementation details}\label{subsec:implementation}
In this section we give the implementation details of all the three \modelName models, \ie, \modelName-M (MoCov2~\cite{chen2020improved}), \modelName-D (DenseCL~\cite{wang2021dense}), and \modelName-S (SimSiam~\cite{chen2021exploring}), respectively. 

\subsubsection{\modelName-M and \modelName-D}
\begin{itemize}
    \item \textbf{Training settings.} To accelerate the training, we train the \modelName-M and \modelName-D with a total batch size of 1,024 and initial learning rate 0.15, which is slightly different from the original setting but does not affect the performance and our conclusion as shown in Sec~\ref{sec:bz}. The other settings are the same as in MoCov2~\cite{chen2020improved} and DenseCL~\cite{wang2021dense}. For example, the input images are first randomly cropped and resized to 224 $\times$ 224 by remaining 20\% $\to$ 100\% regions, which is followed by color jitter with a probability of 0.8, grayscale with a probability of 0.2, Gaussian blur with a probability of 0.5, and random horizontal flips, which are the same as the origin settings in the two base methods. The SGD~\cite{sutskever2013importance} optimizer is adopted with weight decay at 1e-4 and momentum at 0.9.
    \item \textbf{Architecture settings.} There is no difference in the model's architectures comparing the base models and \modelName models. The instance-level and region-level features share the same projector, which has the structure of \textit{Linear(2048, 2048) $\to$ ReLU $\to$ Linear(2048, 128)}, where 2048 is the hidden dimension and 128 is the output dimension. The dense correspondences projectors in DenseCL and \modelName-D have the structure of \textit{Conv2d(2048, 2048) $\to$ ReLU $\to$ Conv2d(2048, 128)}, where the kernel size for the convolutions is $1 \times 1$, and 128 is the output dimension. The length of the memory queue is 65,536.  
\end{itemize}

\subsubsection{\modelName-S}
\begin{itemize}
    \item \textbf{Training settings.} As there is no component like memory queues in SimSiam's architecture, we follow exactly the same training settings as in origin SimSiam to train the \modelName model \modelName-S. Specifically, a total batch size of 512 is employed during the pretraining process. The SGD optimizer with learning rate 0.1, weight decay 1e-4, and momentum 0.9 is adopted to train the model. The data augmentation strategy is the same as the in MoCov2~\cite{chen2020improved} and DenseCL~\cite{wang2021dense}, \ie, the input images are first randomly cropped and resized to 224 $\times$ 224 by remaining 20\% $\to$ 100\% regions, which is followed by color jitter with a probability of 0.8, grayscale with a probability of 0.2, Gaussian blur with a probability of 0.5, and random horizontal flips. The models are trained for 100 epochs with cosine learning rate scheduler. 
    \item \textbf{Architecture settings.} There is no modification on the model's architectures settings comparing the SimSiam and \modelName-S. The projection head follows average pooling or mask pooling and has 3 layers to project the features, which takes the form of $Linear(2048,2048) \to BN(2048) \to ReLU \to Linear(2048,2048) \to BN(2048) \to ReLU \to Linear(2048, 2048) \to BN(2048)$. The predictor head has 2 layers to further process the features and align them with the features extracted from the key views. The structure of the predictor head is $Linear(2048,512) \to BN(512) \to ReLU \to Linear(512,2048)$. These structures are the same as the original settings in SimSiam.
\end{itemize}

\clearpage

\bibliographystyle{splncs04}
\bibliography{egbib}
\end{document}